\pdfoutput=1
\documentclass[letterpaper,journal]{IEEEtran}
\usepackage[T1]{fontenc}
\usepackage[utf8]{inputenc}
\usepackage{amsmath,amsfonts,amssymb,booktabs,graphicx,array,xcolor,url,multirow,etoolbox}
\IfFileExists{placeins.sty}{\usepackage{placeins}}{}
\IfFileExists{stfloats.sty}{\usepackage{stfloats}}{}
\IfFileExists{cite.sty}{\usepackage{cite}}{}
\usepackage[hidelinks,unicode]{hyperref}
\graphicspath{{figures/}}
\renewcommand{\arraystretch}{1.00}
\newcommand{\SBMainTableFont}{%
  \fontsize{7.8}{9.1}\selectfont
  \setlength{\tabcolsep}{3.5pt}%
  \renewcommand{\arraystretch}{1.00}%
}
\newcommand{\SBMainTableNote}{\fontsize{7}{8.5}\selectfont\raggedright}
\newcommand{\SBHeaderStack}[1]{\begin{tabular}[c]{@{}c@{}}#1\end{tabular}}
\makeatletter
\renewcommand{\@IEEEBIOskipN}{32pt}
\expandafter\patchcmd\csname\string\IEEEbiography\endcsname
  {\vskip \@IEEEBIOskipN plus 1fil minus 0\baselineskip}
  {\vskip \@IEEEBIOskipN}
  {}
  {\PackageWarning{ScopeBind}{Biography spacing patch failed}}
\makeatother

\newcommand{\reg}{\mathcal{R}}

\input{experiment-placement}
\begin{document}
\raggedbottom
\title{ScopeBind: Attributing Edited Images to Registered Watermarked Releases}

\author{Hailong~Wang, Wuzheng~Tan, Zixin~Zhang, Xujun~Fang, Fengjuan~Gong,\\%
and~Jian~Weng,~\IEEEmembership{Senior~Member,~IEEE}%
\thanks{This work was supported by the National Natural Science
Foundation of China under Grant 62272199. The work of Jian Weng was
supported in part by the National Natural Science Foundation of China
under Grants 61825203, 62332007, and U22B2028.
(Corresponding author: Wuzheng Tan.)}%
\thanks{Hailong Wang, Wuzheng Tan, Zixin Zhang, Xujun Fang,
and Fengjuan Gong are with the College of Cyber Security,
Jinan University, Guangzhou 510632, China
(e-mail: along12243@163.com; tanwuzheng@jnu.edu.cn;
3432419215@qq.com; 2595349744@qq.com; 2181748278@qq.com).}%
\thanks{Jian Weng is with the School of Cyberspace Security,
Guangzhou University, Guangzhou 510006, China
(e-mail: cryptjweng@gmail.com).}%
}

\markboth{}{}

\maketitle

\begin{abstract}
Attributing an edited image to a registered watermarked release faces a central tension: visual retrieval can match an edited image to its archived original even when it never descended from a watermarked release, whereas generative reconstruction often preserves scene content while erasing readable watermark codes. We present ScopeBind, a registered-release attribution framework that recovers exact release identifiers without a candidate claim at inference. ScopeBind combines global and local watermark hypotheses with a conditional original-gallery proposal, then verifies each candidate via a release-specific orthogonal latent decomposition derived from the registered original--release pair. Projecting the query displacement onto the issuance direction measures alignment with the watermarked release, while the orthogonal remainder retains content-consistency evidence that cancels out of endpoint distances. Using the same representation, a constrained manifold-aligned fusion rule strengthens watermark survivability before registration without retraining the codecs. Across natural and generated domains, ScopeBind reaches up to 97.50\% and 89.38\% exact attribution under strong generative reconstruction, outperforming the strongest public baseline by over 34 percentage points on COCO while keeping transfer and unmarked false acceptances near zero. Cross-editor tests on Stable Diffusion XL and InstructPix2Pix, writer-transfer studies, and gallery scaling to 10,000 originals confirm its robustness and scalability.
\end{abstract}

\begin{IEEEkeywords}
Image provenance, robust watermarking, registered attribution, generative editing, watermark transplantation.
\end{IEEEkeywords}

\section{Introduction}
\IEEEPARstart{C}{onsider} two edited images retrieved from a publisher's archive: one descended from a watermarked release, and the other was reconstructed from the same photograph before watermarking. Although both images match the archived photograph visually, only the first carries the publication lineage that the publisher is asked to confirm. Conversely, if a valid watermark identifier is transplanted onto an unrelated photograph, a conventional decoder can recover every bit accurately and still attribute the image to the wrong release. Distinguishing genuine release lineage from visual similarity or transplanted payloads is essential whenever attribution supports a copyright or provenance claim.

This paper studies attribution to a \emph{registered release}, where a publisher stores each original image alongside its distributed watermarked release and a shared identifier (UID). Given only a single edited query and access to this archive, the receiver must identify the originating release or abstain, without receiving a candidate UID, reference image, crop coordinates, or editing history from the claimant. The framework accommodates both photographs and generated images and focuses on verifying the lineage of the distributed file rather than legal authorship outside the image signal.

Two common transformations expose complementary failure modes in this setting. Spatial cropping removes most of the frame while often leaving a locally readable watermark, whereas generative reconstruction preserves scene composition while suppressing the watermark signal used to rank candidate UIDs~\cite{rombach2022ldm,saberi2024robustness,zhao2024removable}. For our equal-weight COCO releases, for example, the local reader places the true UID within its top-64 shortlist on all 160 center crops, yet succeeds on only a single query under the strongest tested reconstruction. Falling back to pure image similarity whenever decoding fails introduces an opposite failure mode, because an edited unmarked original is itself a close visual match to the archived source.

ScopeBind uses the registered archive in two coordinated ways. First, when code-based recovery abstains, visual retrieval against archived originals can propose a single conditional candidate UID. Second, the latent-space displacement between the registered original and its stored release defines a release-specific direction against which any proposed candidate can be verified. Projecting the query displacement onto this direction measures proximity to the two registered versions, while the orthogonal remainder retains content-consistency evidence that cancels out of the squared-distance difference between the two registered endpoints. Combining both cues enables reliable verification of reconstructed queries while preserving local watermark recovery on cropped images.

Our main contributions are summarized as follows:
\begin{itemize}
\item We formalize attribution to a registered watermarked release via exact UID recovery under two distinct negative populations: edited unmarked originals of the same source and transferred watermark evidence on unrelated images.
\item We derive a release-specific orthogonal latent decomposition from each registered original--release pair and design a single fitted verification head in the recovery branch that tests both issuance alignment and orthogonal content consistency across a bounded set of code and gallery hypotheses.
\item We formulate a constrained finite angle-search fusion rule around this same representation to combine global and local watermark carriers prior to registration under a per-image quantized peak signal-to-noise ratio (PSNR) fidelity bound of 35~dB.
\item We evaluate ScopeBind across two source-disjoint domains, including same-pixel controls, writer-specific verification transfer, frozen cross-editor tests on SDXL and InstructPix2Pix, and nested original-gallery scaling up to 10,000 records.
\end{itemize}

\section{Related Work}
\subsection{Watermark Survival Under Edits}
Learning-based watermarking couples an encoder and decoder through differentiable distortion layers, as exemplified by HiDDeN and StegaStamp~\cite{zhu2018hidden,tancik2020stega}. To address generative image manipulation, Robust-Wide and VINE specifically target instruction-driven and diffusion-based editing~\cite{hu2024robustwide,lu2025vine}, while WaTeRFlow learns from temporal consistency and also provides an image codec~\cite{kuai2026waterflow}. Systematic benchmarks such as WAVES further show that watermark survival depends strongly on both the editor and the writer~\cite{an2024waves}, motivating evaluation across multiple reconstruction strengths on matched sources.

Localized and crop-robust readouts offer a complementary trade-off: WAM provides localized message extraction~\cite{sander2025wam}, while crop-robust methods such as Pixel Seal recover messages from surviving image regions~\cite{soucek2025pixelseal}; meanwhile, EditGuard and OmniGuard jointly address manipulation localization and copyright protection~\cite{zhang2024editguard,zhang2025omniguard}. In ScopeBind, Pixel Seal acts as the local carrier branch to nominate candidates even when only a cropped fragment survives, yet the receiver's decision applies to the query as a whole; a genuine patch pasted onto an unrelated image therefore provides a direct negative control.

\subsection{Positive Attribution and Provenance Retrieval}
Generation-time watermarking methods embed signals directly into a decoder, initial noise realization, or denoising trajectory, as demonstrated by Stable Signature, Tree-Rings, and Gaussian Shading~\cite{fernandez2023stable,wen2023ring,yang2024gaussian}. More recently, Qi \emph{et al.} employ a circular harmonic representation of initial noise to balance distortion against image edits~\cite{qi2026harmonic}. Rather than measuring partial bit survival against a known expected message, our setting requires selecting one exact UID from a 16-bit code space without a supplied claim, including for photographs that have no generation trajectory.

Retrieving an archived record via watermarks or perceptual fingerprints is a well-established mechanism in provenance standards and soft-binding frameworks~\cite{c2pa24guidance,kazempourradi2026soft}. Copy-detection descriptors such as SSCD and general visual backbones such as DINOv2~\cite{pizzi2022sscd,oquab2024dino} remain informative when generative edits preserve scene identity while erasing watermark codes, but visual retrieval alone cannot distinguish a watermarked release from its unmarked original.

\subsection{Content-Bound Watermarking Against Transplantation}
Watermark transplantation attacks demonstrate that accepting a decoded payload in isolation is vulnerable to cross-image copying~\cite{kutter2000copy}. To counter this, SEAL binds watermark keys directly to semantic image content~\cite{arabi2025seal}. Closely related to our verification problem, TetherMark checks both a message and a registered patch relation to authenticate a claimed instance under transplantation~\cite{wei2026tether}. ScopeBind must also identify the candidate record blindly; once a record is proposed, its registered original--release pair defines an explicit release direction so that the receiver can verify whether the query supports the watermarked release even when the unmarked original is a near-identical visual match. Our experiments evaluate the selected watermarking baselines under a common UID interface and isolate the contribution of access to the registered release record.

\section{Proposed Method}
\subsection{Problem Formulation and Threat Model}
The publisher assigns each source a unique identifier $u\in\{0,\ldots,2^{16}-1\}$ and enrolls both the original image $x_u$ and its distributed 8-bit watermarked release $w_u$ in a trusted registry $\reg$, which may also include unissued original-only distractors. Given an edited RGB query $y$, the receiver outputs either a single UID or an abstention ($\bot$). Edited versions of the distributed release $T(w_u)$ are positive queries for $u$; edited versions of the corresponding unmarked original $T(x_u)$ are negative queries even when their visual scene remains recognizable. Additional negative controls place a valid UID, a release residual, or pixels copied from an issued image onto an unrelated carrier, including cases where an attacker has access to a writer capable of reproducing the target payload; the protocol therefore does not assume that public codewords are secret or uncopyable.

\begin{figure*}[!t]
\centering
\includegraphics[width=.96\textwidth]{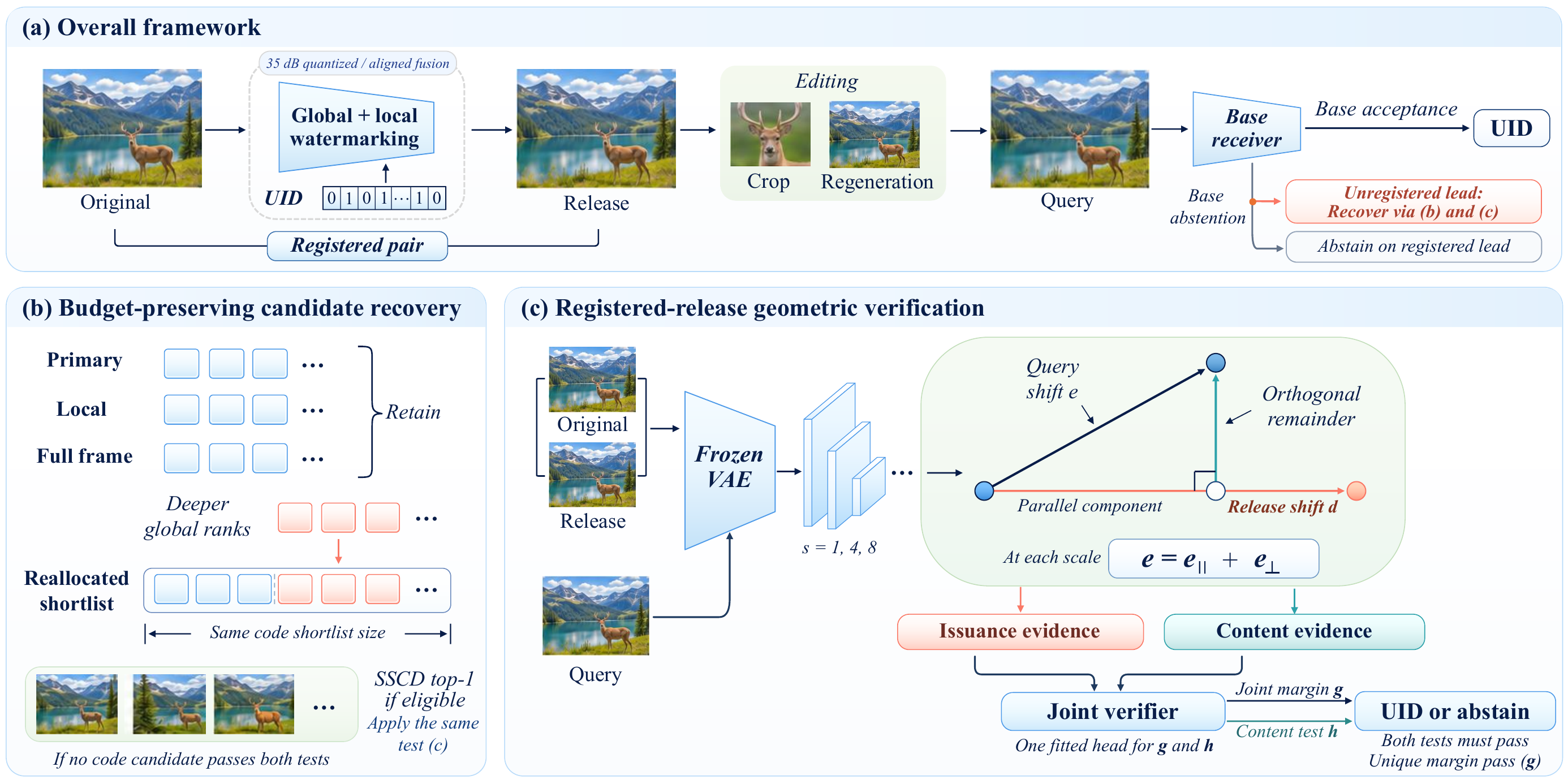}
\caption{Overview of the ScopeBind framework. (a)~Overall framework and base routing: global and local watermarking embed a shared UID under a 35~dB quantized PSNR floor to form a registered original--release pair; edited queries that abstain at the base receiver with an unregistered leading UID enter recovery, whereas registered-lead abstentions are preserved. (b)~Budget-preserving candidate recovery: primary, local, and full-frame candidates are retained while vacated slots are filled from deeper global ranks under a fixed shortlist budget, with one conditional top-1 SSCD gallery proposal tested only if no code candidate passes. (c)~Registered-release geometric verification: a frozen VAE encodes the registered pair and query across scales $s\in\{1,4,8\}$, decomposing the query displacement into a parallel issuance component and an orthogonal content remainder ($e = e_\parallel + e_\perp$, corresponding to $e_s = a_s\hat{d}_s + v_s$ in~\eqref{eq:projection}); a single fitted head evaluates both the joint margin $g$ and content test $h$ to accept a supported UID or abstain.}
\label{fig:pipeline}
\end{figure*}

Figure~\ref{fig:pipeline} traces an image from publication to attribution. Panel~(a) illustrates dual-carrier registration, editing, and base-receiver routing, where abstentions with an unregistered leading UID enter recovery while registered-lead abstentions are preserved. Panel~(b) summarizes budget-preserving candidate recovery and the conditional top-1 SSCD gallery proposal. Panel~(c) depicts multiscale latent decomposition in the frozen VAE space and the shared joint verifier that applies both the joint margin and orthogonal content tests. Experimental queries and attribution decisions are evaluated in Section~\ref{sec:experiments}.

\subsection{Dual-Carrier Publishing and Manifold-Aligned Fusion}
VINE and Pixel Seal embed the same logical UID through their native global and local interfaces~\cite{lu2025vine,soucek2025pixelseal}. Specifically, the global branch embeds a masked, shortened 100-bit BCH frame at $256\times256$ and resizes the resulting residual, and the local branch embeds those same bits alongside 156 fixed padding bits. On the prepared $512\times512$ original, combining the two adjusted RGB residuals $\delta_g$ and $\delta_l$ yields the quantized equal-weight release
\begin{equation}
w_u=Q_8\!\left(\operatorname{clip}\bigl(x_u+\alpha(\delta_g+\delta_l),0,1\bigr)\right),
\label{eq:publish}
\end{equation}
where the scalar strength $\alpha>0$ is searched along this residual direction until the stored 8-bit image satisfies a 35~dB PSNR floor. Storing the exact quantized release ensures that downstream verification measures query displacements directly against the clipped and quantized pixels actually distributed.

While~\eqref{eq:publish} provides a fixed equal-weight baseline $w_u^{\mathrm{EW}}$, we also formulate a release-manifold-aligned writing rule guided by the same frozen verification VAE encoder $E(\cdot)$ used at inference, aligning the fused residual with the VAE reconstruction manifold. Let $P_s$ denote nonoverlapping $s\times s$ spatial block averaging, and let $\bar\delta_g$ and $\bar\delta_l$ denote the two residuals normalized by their RGB root mean square. For each angle $\theta$ on a fixed 17-point grid over $[0.15\pi,0.35\pi]$, we define
\begin{equation}
 w_u(\theta)=Q_8\!\left(\operatorname{clip}\bigl(x_u+\rho_\theta(
 \cos\theta\,\bar\delta_g+\sin\theta\,\bar\delta_l),0,1\bigr)\right),
 \label{eq:aligned_write}
\end{equation}
where $\rho_\theta>0$ is determined via binary search on the quantized 8-bit image so that every candidate satisfies the 35~dB PSNR constraint. Including the equal-weight baseline alongside the 17 angle-grid points forms an 18-candidate feasible set while leaving both native writers and codebooks frozen.

For each candidate release, let $d_s=P_s(E(w_u)-E(x_u))$, $\hat d_s=d_s/\|d_s\|_2$, and let $\mathcal R_{\mathrm{VAE}}$ denote deterministic reconstruction through the frozen verification VAE. With $\epsilon=10^{-8}$, the writing objective is
\begin{align}
 J(w_u)={}&\sum_{s\in\{1,4\}}
 \frac{\langle P_s(E(\mathcal R_{\mathrm{VAE}}(w_u))-E(x_u)),\hat d_s\rangle}
 {\|d_s\|_2+\epsilon}\nonumber\\
 &+\mathcal M_g(w_u,u)+\mathcal M_l(w_u,u),
 \label{eq:aligned_objective}
\end{align}
where each margin $\mathcal M$ equals the issuing codeword score minus the highest score of any other public UID, normalized by the 100 payload bits, with unit weighting. Selecting the feasible candidate $w_u^*$ that maximizes $J(w_u)$ couples writing and verification through the actual registered displacement without training an additional network or observing any edited test query, guaranteeing $J(w_u^*)\ge J(w_u^{\mathrm{EW}})$ under the per-image 35~dB quantized PSNR floor. Because $J(w_u)$ optimizes a deterministic VAE reconstruction surrogate rather than a specific downstream diffusion editor, we retain both equal-weight and manifold-aligned releases in Section~\ref{sec:experiments} to evaluate their paired robustness, cross-editor generalization, false-acceptance rates, and fidelity.

\subsection{Base Routing and Budget-Preserving Candidate Recovery}
At inference, each reader ranks the 65,536 public codewords across 19 fixed views (the full image, 12 bicubic rotations from $\pm 5^\circ$ to $\pm 30^\circ$ in $5^\circ$ steps, three quarter rotations, and three centered reflect-padded scalings at 0.85, 0.90, and 0.95); after undoing the public bit mask, the score of codeword $c(u)$ is $\ell^\top c(u)$ for the view's bit log-odds $\ell$. The two-stage base receiver $B(y)$ represents each candidate with 116 features (16 code, 24 registered regional, and 76 cross-view rank features) scored by frozen three-layer GELU heads over the top-16 primary and remaining top-64 tail candidates per view (architecture and frozen thresholds are detailed in Section~\ref{sec:experiments}). ScopeBind returns any acceptance by $B(y)$ immediately. If $B(y)$ abstains while the leading UID of the untransformed global view already has a registry entry, the receiver preserves that abstention so that a rejected leading claim cannot be reassigned during fallback.

For the remaining abstentions, let $C_0$ denote the deduplicated union of the top-64 lists. ScopeBind retains all primary candidates, all local top-64 lists, and the full untransformed global top-64 list, then fills the vacated slots with the highest-ranked UIDs outside $C_0$ from the untransformed global view to form $C(y)$ with $|C(y)|=|C_0|$. This reallocation preserves the distinct code-candidate budget per query, while conditional retrieval adds at most one further candidate; only records with an enrolled release enter geometric scoring.

\subsection{Release-Specific Orthogonal Latent Decomposition}
For each candidate record, the frozen VAE encoder $E$ computes the posterior mean of the query once per verification call and reuses the cached posterior means of the registered pair. Using the block-averaging operator $P_s$ defined above, for any nonzero release displacement $d_s$ we decompose the query displacement $e_s$ into a parallel component along the release direction and an orthogonal remainder:
\begin{align}
 d_s&=P_s(E(w_u)-E(x_u)), & e_s&=P_s(E(y)-E(x_u)),\nonumber\\
 \hat d_s&=d_s/\|d_s\|_2, & a_s&=\langle e_s,\hat d_s\rangle,\nonumber\\
 v_s&=e_s-a_s\hat d_s, & \langle v_s,d_s\rangle&=0,
 \label{eq:projection}
\end{align}
where inner products sum over all channels and spatial positions. As illustrated in Fig.~\ref{fig:pipeline}(c), the origin represents the registered original, the horizontal endpoint represents its stored release, and the foot of the perpendicular gives the projection $a_s\hat d_s$. Subtracting the squared distances from the query to the two registered endpoints yields the identity
\begin{equation}
 \|e_s\|_2^2-\|e_s-d_s\|_2^2
 =2\|d_s\|_2\bigl(a_s-\|d_s\|_2/2\bigr).
 \label{eq:midpoint}
\end{equation}
Although the sign of~\eqref{eq:midpoint} determines which side of the perpendicular bisector contains the query, this squared-distance difference cancels $v_s$ completely. Consequently, a query can appear closer to the release than to the original even under a large orthogonal content displacement, whereas adding $\beta d_s$ to $e_s$ shifts the projection while leaving $v_s$ invariant. Retaining both components therefore separates issuance alignment from content consistency in latent space.

\subsection{Shared Joint Verifier and Dual Acceptance Tests}
To summarize this decomposition compactly, we retain the centered projection $a_s-\|d_s\|_2/2$ and release norm $\|d_s\|_2$ at scales $s\in\{1,4\}$. At scales $s\in\{1,4,8\}$, we partition $v_s$ into a $4\times4$ spatial grid, compute the root mean square within each cell, and apply $\log(1+q)$ to the lower and upper quartiles $q$ of the 16 cell values, yielding a 10-dimensional feature vector $\phi$ consisting of four projection-related entries and six orthogonal-remainder entries. A single class-balanced linear softmax head is fitted over three classes: issued ($I$), same-source unmarked ($U$), and invalid claim ($X$), producing class logits $L_c(\phi)$ for $c\in\{I,U,X\}$ from standardized features with weights $W_{j,c}$ and training feature scales $\sigma_j$. With remainder index set $\mathcal O$, the joint margin and content-consistency score are defined as
\begin{align}
g(\phi)&=L_I(\phi)-\max\{L_U(\phi),L_X(\phi)\},\label{eq:joint}\\
h(\phi)&=\sum_{j\in\mathcal O}(W_{j,I}-W_{j,X})\phi_j/\sigma_j.
\label{eq:witness}
\end{align}
Here $h(\phi)$ combines the six remainder features scaled by $\sigma_j$ with the issued-versus-invalid weights of the same head (without intercept, as any per-feature centering constant is absorbed into the threshold $\kappa$), enforcing explicit support from the orthogonal component that~\eqref{eq:midpoint} discards. The recovery head is fitted on an earlier source split, with the joint cutoff $\tau$ calibrated on a disjoint negative split and the content cutoff $\kappa$ set from the fitting positive claims; the resulting frozen operating point and split sizes are detailed in Section~\ref{sec:experiments}.

\subsection{Decision Rule and Conditional Gallery Proposal}
Among scored code candidates, let $J$ be the subset satisfying $g>\tau$ and $H\subseteq J$ the subset also satisfying $h\ge\kappa$. The receiver accepts the member of $J$ only when it is unique and belongs to $H$; if $H$ is nonempty while $J$ contains competing candidates, it abstains. When $H$ is empty, the receiver ranks \emph{all} registered originals by cosine similarity between normalized SSCD descriptors of $y$ and $x_u$~\cite{pizzi2022sscd} and tests the top-1 match only if its release is registered and its UID lies outside $C(y)$. An ineligible top-1 match terminates the search; an eligible match is accepted only if it passes both thresholds $\tau$ and $\kappa$. This single-proposal gate ensures that visual retrieval only nominates a plausible source, while acceptance still requires geometric evidence supporting the issued release.

\section{Experiments}\label{sec:experiments}

\subsection{Experimental Setup}

We evaluate ScopeBind on two source-disjoint galleries, each comprising 160 issued images and 96 original-only distractors drawn from COCO test2017 photographs~\cite{lin2014coco} and DiffusionDB generated images~\cite{wang2023diffusiondb}, respectively. Both test sets are strictly disjoint from the training and calibration splits, so no source or prompt overlaps with parameter fitting or threshold calibration.

Each source produces 13 legitimate queries spanning the clean release, three resized crops (75\% center, 75\% upper-left, and 60\% lower-right side length), JPEG compression (quality 50), two rotations ($17^\circ$ and $30^\circ$), deterministic VAE reconstruction, three image-to-image reconstructions (strengths 0.3, 0.4, and 0.5), and two inpaintings ($128\times128$ and $256\times256$). Image-to-image editing uses Stable Diffusion 2.1-base (SD2.1) with an empty prompt and PNDM~\cite{liu2022pndm}, with the strong reconstruction endpoint using strength 0.5 on a 50-step schedule. Applying the same 13 conditions to unmarked originals tests whether visual identity alone triggers false acceptance. Seven additional copy controls per source test watermark transplantation and patch splicing onto an unrelated donor: direct target-UID writing on the donor (full image and $384\times384$ upper-left crop), 35~dB release-residual transfer onto the clean donor, and four pixel-splice composites replacing a $256\times256$ center square, a $384\times384$ center square, the right half, or offset $128\times128$ checkerboard tiles, totaling 2,080 positive, 2,080 unmarked, and 1,120 transfer queries per domain.

All baseline writers (Robust-Wide, VINE, Pixel Seal, WAM, and the WaTeRFlow image codec) share the same 16-bit logical UID, 19 fixed views, and regional K64 verification adapter (where K64 denotes the per-view top-64 prefix scored with a shared calibrated regional threshold), producing separate releases at the common 35~dB PSNR floor with matched sources, edits, and random seeds. Dual + K64 reads the exact same dual-carrier release pixels as ScopeBind using only this shared regional K64 adapter, without ScopeBind's learned two-stage base receiver $B(y)$ or geometric fallback; by contrast, the no-gallery control holds the full ScopeBind receiver and both thresholds fixed while disabling conditional retrieval.

A positive query is counted as correct only when the returned UID matches the issuing UID exactly, with wrong-UID outputs and abstentions tracked separately, and any accepted UID on a negative query counted as a false acceptance (FA). Issuance fidelity is measured against the original via PSNR and SSIM~\cite{wang2004ssim}. All paired differences are computed within domain, with 95\% percentile confidence intervals obtained from 10,000 source-level bootstrap resamples that keep all conditions from each source grouped. For the ScopeBind receiver, the two-stage base heads adopt $116\text{--}32\text{--}3$ GELU architectures (primary threshold 4.72 and tail threshold 0), while the linear recovery head is fitted on 96 training sources and calibrated on a disjoint 32-source validation split, yielding $\tau=2.35$ and $\kappa=4.73$.
\switchonecolumn
\par\smallskip
{%
  \SBTablePanel{\RenewDocumentEnvironment{table*}{O{}}{}{}\centering
\caption{Attribution of edited RGB queries across two held-out image domains.}
\label{tab:comparison}
\SBMainTableFont

\begin{tabular*}{\linewidth}{@{\extracolsep{\fill}}l*{10}{c}@{}}
\toprule
\multirow[c]{2}{*}[-0.6ex]{Method} & \multicolumn{5}{c}{COCO ($n=160$)}
& \multicolumn{5}{c}{DiffusionDB ($n=160$)}\\
\cmidrule(lr){2-6}\cmidrule(lr){7-11}
& Strong$\uparrow$ & All$\uparrow$ & Crop$\uparrow$ & Copy FA$\downarrow$ & Unmarked FA$\downarrow$ & Strong$\uparrow$ & All$\uparrow$ & Crop$\uparrow$ & Copy FA$\downarrow$ & Unmarked FA$\downarrow$\\
\midrule
\textbf{ScopeBind} & \textbf{95.63} & \textbf{98.46} & 98.33 & \textbf{0.00} & \textbf{0.00} & \textbf{86.88} & \textbf{97.45} & 96.67 & \textbf{0.09} & \textbf{0.10} \\
Robust-Wide + K64 & \underline{61.25} & 66.01 & 2.29 & 1.61 & 2.50 & \underline{60.00} & 64.52 & 1.88 & 0.63 & 1.78 \\
VINE + K64 & 60.63 & 65.72 & 2.08 & 1.25 & 2.16 & 55.00 & 64.57 & 2.08 & 0.98 & 1.44 \\
Pixel Seal + K64 & 1.25 & 76.11 & \underline{99.58} & \underline{0.27} & 1.59 & 0.00 & 74.33 & \textbf{98.54} & 0.45 & 1.30 \\
WAM + K64 & 0.00 & 76.54 & \textbf{99.79} & 0.36 & \underline{0.14} & 0.00 & 72.21 & \underline{97.92} & \underline{0.18} & \underline{0.34} \\
WaTeRFlow + K64 & 56.88 & 79.57 & 31.88 & 1.07 & 0.77 & 51.25 & 77.31 & 29.79 & 1.16 & 2.12 \\
\addlinespace[2pt]
Dual + K64 & 56.88 & \underline{95.38} & 99.38 & 1.07 & 3.75 & 53.13 & \underline{93.70} & 97.29 & 1.96 & 2.74 \\
\bottomrule

\end{tabular*}
\par\smallskip
{\SBMainTableNote
Rates are percentages. Strong denotes correct attribution under PNDM 0.5 ($160$ queries per domain). All and Crop aggregate $2{,}080$ and $480$ positive queries. Copy FA and Unmarked FA use $1{,}120$ and $2{,}080$ negative queries. Each comparator uses regional K64. The Dual control shares ScopeBind's issued pixels. All systems make zero wrong-UID decisions on positive queries. Bold and underlining mark the best and second-best distinct scores.\par}
}}

\subsection{Attribution Under Strong Generative Reconstruction}

\ScopeColumns{
Under the strongest tested reconstruction (PNDM 0.5), the equal-weight ScopeBind release correctly attributes 153 of 160 COCO queries (95.63\%) and 139 of 160 DiffusionDB queries (86.88\%, Table~\ref{tab:comparison}). By comparison, Robust-Wide + K64, the strongest public baseline at this endpoint, identifies 98 COCO and 96 DiffusionDB sources, yielding gains of 34.38 percentage points on COCO (95\% confidence interval 26.25 to 42.50) and 26.88 percentage points on DiffusionDB (18.75 to 35.00). Source-paired decision counts clarify how these gains arise: relative to Robust-Wide + K64 on COCO, ScopeBind matches 93 recoveries, uniquely recovers 60 additional sources, and misses 5 that the baseline recovers (with 2 missed by both); on DiffusionDB, it matches 89 recoveries, uniquely recovers 50 additional sources, and misses 7 (with 14 missed by both).
}{
More directly, when both receivers operate on identical dual-carrier pixels, ScopeBind matches 90 recoveries of Dual + K64 on COCO and 84 on DiffusionDB, uniquely recovers 63 missed sources on COCO and 55 on DiffusionDB, and loses only a single source in each domain (with 6 COCO and 20 DiffusionDB sources missed by both); this confirms substantial receiver-side gains under identical issued pixels. Relative to the gallery-free ScopeBind control, conditional gallery retrieval adds 17 strong recoveries on COCO (136 recovered by both, 17 uniquely by gallery retrieval, 0 lost, and 7 missed by both) and 12 on DiffusionDB (127 recovered by both, 12 uniquely by gallery retrieval, 0 lost, and 21 missed by both). Because evaluation requires exact identifier recovery under the shared adapter, partially recovered messages receive no credit.
}

\par\smallskip
\ScopeColumns{
Figure~\ref{fig:qual_strong} illustrates representative outcomes across all four behavioral regimes. On COCO, the cowshed case (UID~48302) is recovered by gallery-assisted ScopeBind and Robust-Wide while Dual abstains, and the dog case (UID~21104) is recovered by ScopeBind along the code-verification path despite Dual abstaining on identical pixels.
}{
On DiffusionDB, the waterfall scene (UID~37158) is recovered uniquely by ScopeBind while every compared baseline abstains. Conversely, all evaluated receivers abstain on the woodland pagoda scene (UID~36775), which shows that visual recognizability alone does not suffice to certify release lineage when the evaluated receivers cannot recover sufficient evidence for attribution.
}

\par\smallskip
\SBPanel{\centering
\includegraphics[width=.75\linewidth]{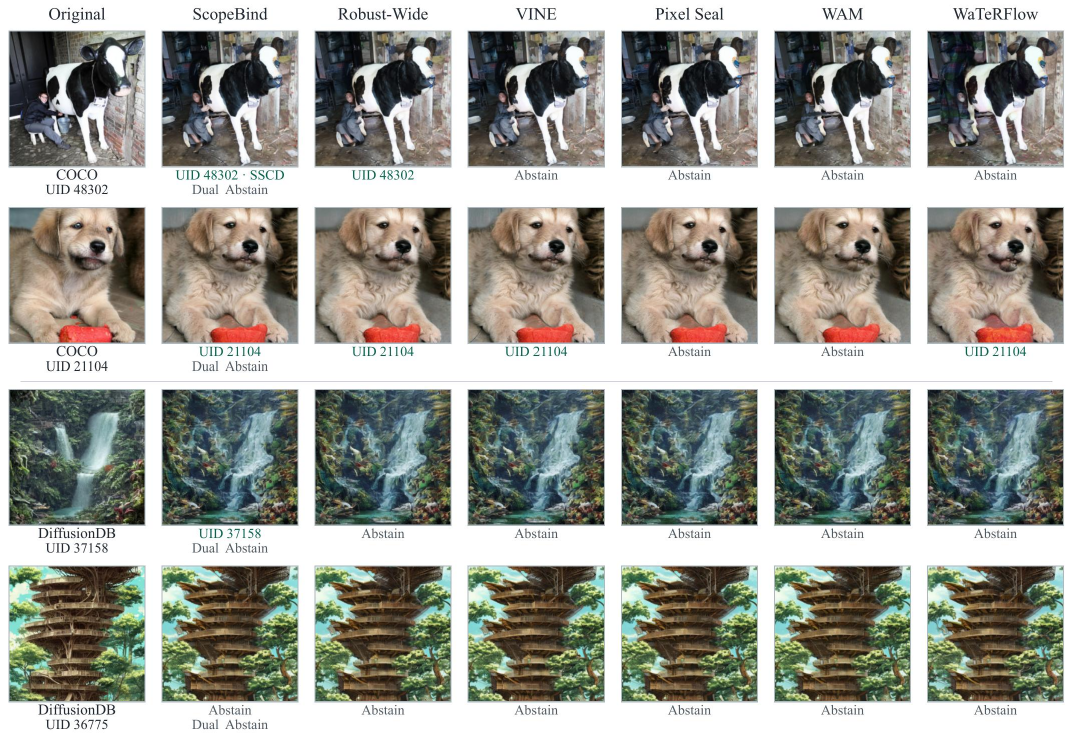}
\caption{Representative COCO (top two rows) and DiffusionDB (bottom two rows) reconstructions at PNDM strength 0.5 with matched sources and random seeds. Verdicts correspond to the displayed queries, where Dual and ScopeBind read identical release pixels.}\label{fig:qual_strong}}
\par\smallskip

\subsection{Robustness Across Image Transformations}
\ScopeColumns{
Aggregated across all 13 conditions, ScopeBind returns 2,048/2,080 correct UIDs on COCO (98.46\%) and 2,027/2,080 on DiffusionDB (97.45\%), outperforming the strongest aggregate baseline Dual + K64 (95.38\% and 93.70\%) by 3.08 and 3.75 percentage points. At PNDM 0.4, ScopeBind already attributes 160/160 COCO and 158/160 DiffusionDB sources correctly, and the separation between receivers widens sharply at strength 0.5 (Fig.~\ref{fig:conditions}).
}{
The same-pixel control reveals how receiver performance scales with reconstruction strength: at PNDM 0.3, ScopeBind and Dual + K64 recover 160 versus 158 COCO sources and 159 versus 156 DiffusionDB sources, but at PNDM 0.5 those margins expand to 62 and 54 sources. Holding sources, releases, and thresholds constant across all three strengths isolates this sensitivity to reconstruction severity. Because spatial cropping instead removes most of the frame while preserving a local patch, we next examine local candidate coverage separately.
}

\par\smallskip
\SBPanel{\centering
\includegraphics[width=.74\linewidth]{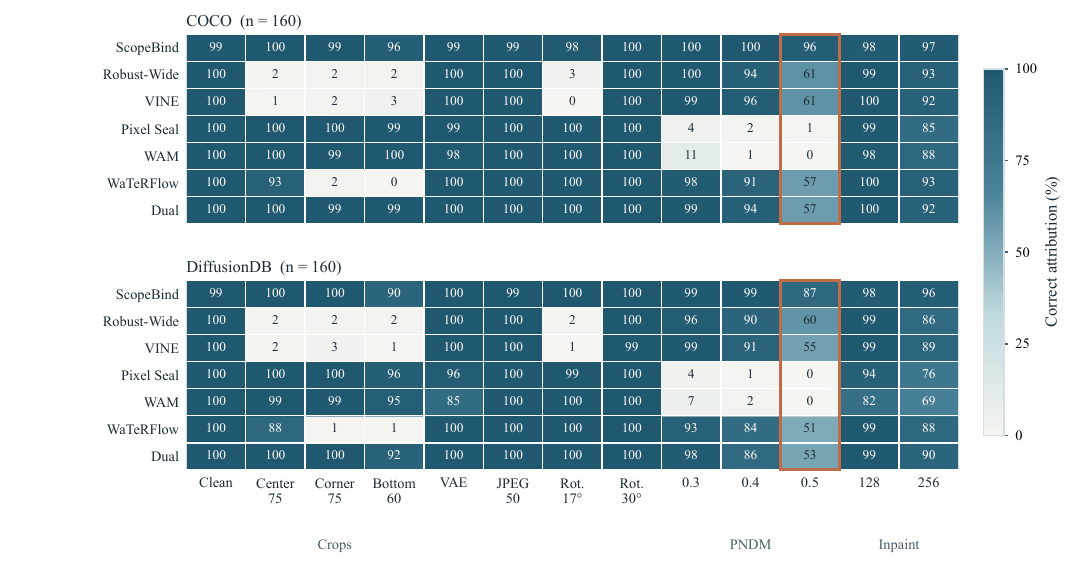}
\caption{Exact attribution rates (\%) across all 13 editing conditions on 160 sources per domain (labels rounded to the nearest integer), with the outlined column highlighting the primary PNDM 0.5 endpoint.}\label{fig:conditions}}
\par\smallskip

\SBPanel{\centering
\includegraphics[width=.84\linewidth]{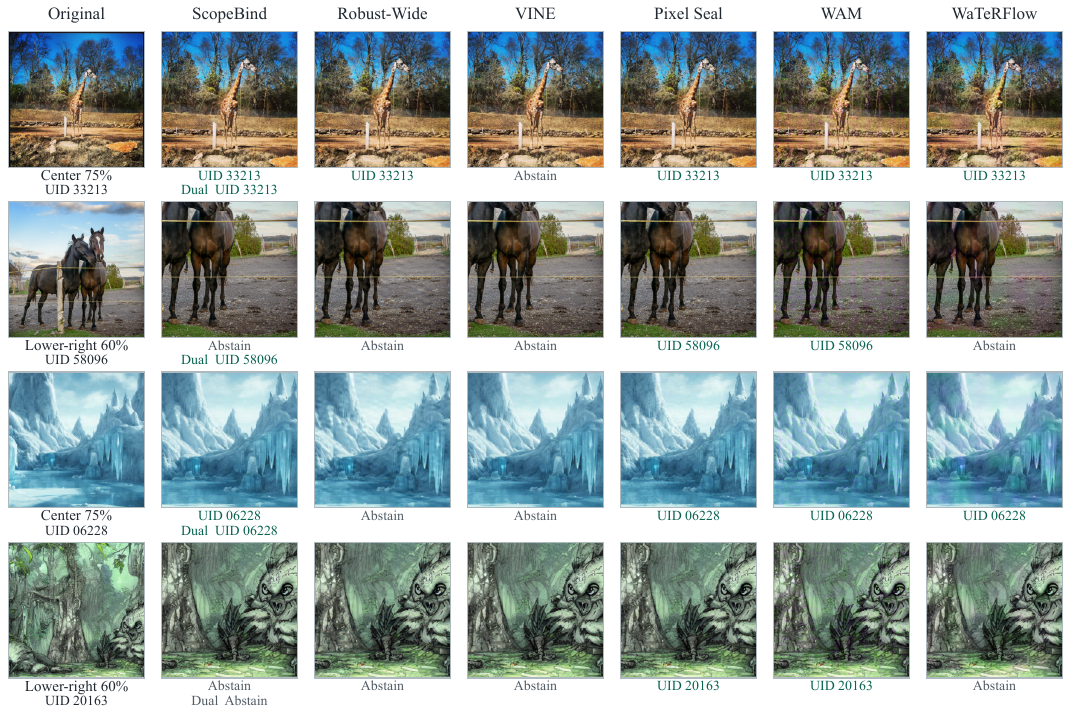}
\caption{Representative cropping outcomes on resized center crops (75\% retained side length) and tighter lower-right crops (60\% retained side length).}\label{fig:qual_crop}}
\par\smallskip
{%
  \SBTablePanel{\RenewDocumentEnvironment{table*}{O{}}{}{}\centering
\caption{True-UID candidate coverage and exact attribution on identical dual-carrier releases.}
\label{tab:nomination}
\SBMainTableFont
\begin{tabular*}{\linewidth}{@{\extracolsep{\fill}}l*{12}{c}@{}}
\toprule
\multirow[c]{2}{*}[-0.6ex]{Condition}
  & \multicolumn{6}{c}{COCO ($n=160$)}
  & \multicolumn{6}{c}{DiffusionDB ($n=160$)}\\
\cmidrule(lr){2-7}\cmidrule(lr){8-13}
  & Global & Local & \SBHeaderStack{Top-64\\union} & \SBHeaderStack{Reallocated\\list}
  & \SBHeaderStack{No\\gallery} & Full
  & Global & Local & \SBHeaderStack{Top-64\\union} & \SBHeaderStack{Reallocated\\list}
  & \SBHeaderStack{No\\gallery} & Full\\
\midrule
Center crop 75 & 3 & 160 & 160 & 160 & 160 & 160 & 2 & 160 & 160 & 160 & 160 & 160 \\
Upper-left crop 75 & 2 & 160 & 160 & 160 & 159 & 159 & 5 & 160 & 160 & 160 & 160 & 160 \\
Lower-right crop 60 & 4 & 158 & 158 & 158 & 153 & 153 & 4 & 154 & 154 & 154 & 144 & 144 \\
PNDM 0.3 & 158 & 2 & 158 & 160 & 160 & 160 & 156 & 3 & 156 & 159 & 158 & 159 \\
PNDM 0.4 & 152 & 3 & 152 & 157 & 157 & 160 & 144 & 2 & 144 & 155 & 153 & 158 \\
PNDM 0.5 & 122 & 1 & 123 & 141 & 136 & 153 & 110 & 2 & 111 & 140 & 127 & 139 \\
\bottomrule

\end{tabular*}
\par\smallskip
{\SBMainTableNote
The first four columns in each domain count true UIDs reaching a candidate list. Global and Local each
count membership in the branch's top-64 in any of 19 views, and Top-64 union counts membership in
either branch. The last two columns count exact final decisions. Crop percentages denote retained side length.\par}
}}

\ScopeColumns{
Across the three cropping conditions, ScopeBind recovers 472/480 COCO and 464/480 DiffusionDB queries, closely tracking the dedicated local baselines WAM (479 on COCO) and Pixel Seal (473 on DiffusionDB). Seven of the eight COCO misses and all sixteen DiffusionDB misses occur on the tighter 60\% lower-right crop (Fig.~\ref{fig:qual_crop}). Inspecting candidate rankings explains why the local branch is essential: on 469 COCO and 463 DiffusionDB crop queries, the true UID enters a local top-64 shortlist while missing every global top-64 list. For the $128\times128$ and $256\times256$ inpainting masks, ScopeBind recovers 156 and 155 COCO queries and 156 and 153 DiffusionDB queries, respectively.
}{
Table~\ref{tab:nomination} traces candidate coverage from initial ranking to final acceptance on identical dual-carrier releases. Reallocating vacated code slots lifts strong-query coverage from 123 to 141 COCO sources and from 111 to 140 DiffusionDB sources before gallery retrieval is ever invoked.
}

\subsection{False-Acceptance Control and Publication Fidelity}

\ScopeColumns{
To verify whether higher positive recovery increases unauthorized claims, we evaluate false acceptances across all negative controls (Table~\ref{tab:comparison}). ScopeBind produces zero wrong-UID decisions across all 4,160 positive queries and accepts only 1 of 2,240 transfer queries (0.00\% on COCO, 0.09\% on DiffusionDB) and 2 of 4,160 unmarked queries (0.00\% on COCO, 0.10\% on DiffusionDB), compared with 12 COCO and 22 DiffusionDB transfer false acceptances plus 78 COCO and 57 DiffusionDB unmarked false acceptances for Dual + K64 on identical pixels.
}{
All three false acceptances occur on DiffusionDB (a donor-cropped patch, a center-cropped unmarked original, and a rotated unmarked original), where the latter two match the visual source yet erroneously certify issuance. On COCO, observing zero false acceptances across 160 independent sources corresponds to a 95\% upper bound of 1.85\%. Because all receiver ablations operate on identical release pixels, fidelity differences apply only across writers: ScopeBind maintains a mean SSIM above 0.986 in both domains at a mean PSNR just above 35~dB (Table~\ref{tab:quality}, Fig.~\ref{fig:qual_fidelity}), while Pixel Seal attains the highest mean PSNR and VINE the highest mean SSIM.
}
\par\smallskip
{%
  \SBTablePanel{\RenewDocumentEnvironment{table*}{O{}}{}{}\begin{table*}[tp]
\centering
\caption{Fidelity of the quantized publication relative to its original.}
\label{tab:quality}
\SBMainTableFont

\begin{tabular*}{\linewidth}{@{\extracolsep{\fill}}l*{6}{c}@{}}
\toprule
\multirow[c]{2}{*}[-0.6ex]{Publication} & \multicolumn{3}{c}{COCO ($n=160$)}
& \multicolumn{3}{c}{DiffusionDB ($n=160$)}\\
\cmidrule(lr){2-4}\cmidrule(lr){5-7}
& Mean PSNR$\uparrow$ & Min. PSNR & SSIM$\uparrow$ & Mean PSNR$\uparrow$ & Min. PSNR & SSIM$\uparrow$\\
\midrule
ScopeBind / Dual & 35.0558 & 35.0000 & \underline{0.98693} & 35.0392 & 35.0000 & \underline{0.98832} \\
Robust-Wide & \underline{35.1984} & 35.0001 & 0.96770 & \underline{35.2580} & 35.0013 & 0.96909 \\
VINE & 35.0192 & 35.0000 & \textbf{0.99353} & 35.0239 & 35.0000 & \textbf{0.99314} \\
Pixel Seal & \textbf{35.3011} & 35.0003 & 0.95473 & \textbf{35.3493} & 35.0003 & 0.96058 \\
WAM & 35.1337 & 35.0005 & 0.95770 & 35.1991 & 35.0003 & 0.96251 \\
WaTeRFlow & 35.1399 & 35.0014 & 0.98472 & 35.1787 & 35.0003 & 0.98707 \\
\bottomrule

\end{tabular*}
\par\smallskip
{\SBMainTableNote
PSNR is in dB. Means are over 160 issued images per cell. All measured minima are at least 35~dB before rounding for display. ScopeBind and Dual use identical publications. Bold and underlining rank the means.\par}
\end{table*}
}}
\par\smallskip
\SBPanel{\centering
\includegraphics[width=.64\linewidth]{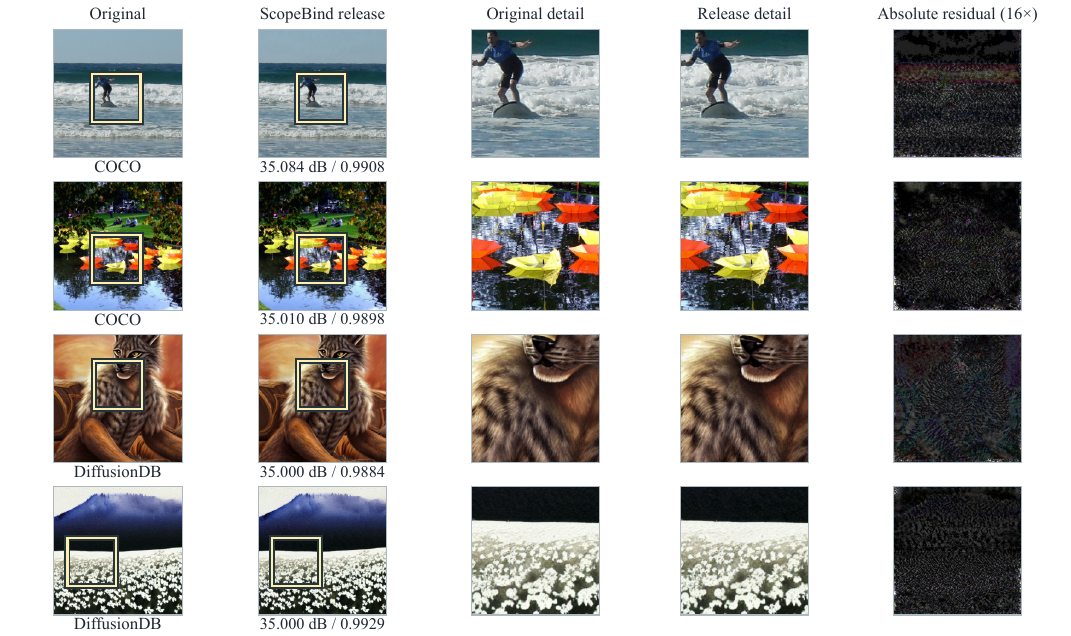}
\caption{Original and distributed releases prior to editing, showing matched detail crops alongside full-frame absolute residuals amplified $16\times$ for visibility.}\label{fig:qual_fidelity}}
\par\smallskip

\SBNeedspace{120pt}
\subsection{Ablation Studies}
\ScopeColumns{Table~\ref{tab:ablation} and Fig.~\ref{fig:stage_results} isolate each component of the decision rule on identical ScopeBind releases, starting from \emph{Original receiver}, which corresponds to ScopeBind's learned two-stage base receiver $B(y)$ prior to geometric fallback (recovering 119 COCO and 105 DiffusionDB strong queries with zero false acceptances, compared with 91 and 85 for the single-threshold Dual + K64 adapter in Table~\ref{tab:comparison}). Enabling joint geometric verification over reallocated candidates raises strong-reconstruction attribution from 119 to 136 on COCO and from 105 to 128 on DiffusionDB, but admits 9 and 7 transfer claims, respectively. Preserving registered-lead abstentions eliminates all transfer false acceptances, leaving 12 unmarked false acceptances and two wrong-UID outputs on COCO. Enforcing the orthogonal remainder test $h\ge\kappa$ removes all 14 remaining errors while costing only one DiffusionDB crop, yielding a gallery-free receiver that recovers 136 COCO and 127 DiffusionDB strong queries with zero observed false acceptances in either domain. Adding the single conditional SSCD gallery proposal recovers 17 further strong queries on COCO (+10.63 points, 95\% interval 6.25 to 15.63) and 12 on DiffusionDB (+7.50 points, 3.75 to 11.88), whereas spending that same one-candidate budget on the next}{watermark-ranked code adds only three and one queries, respectively. Requiring additional agreement with DINOv2 retains 150 COCO and 139 DiffusionDB strong recoveries without reducing the three observed DiffusionDB false acceptances. All gallery-stage variants preserve the same crop attribution counts (472 on COCO and 464 on DiffusionDB), and every negative acceptance in the final system enters through the gallery, making both the gain and cost of retrieval directly measurable. Although conditional retrieval brings the true UID within reach on all 320 strong queries, 28 still abstain: 25 fall below the joint margin cutoff, two have competing code candidates above cutoff, and one is blocked by the registered-lead rule. In addition, the highest negative gallery margin on DiffusionDB (3.59) exceeds the median accepted true gallery margin (3.13), indicating that raising a single global threshold to eliminate that error would also reject valid recoveries.}

\par\smallskip
{%
  \SBTablePanel{\RenewDocumentEnvironment{table*}{O{}}{}{}\begin{table*}[tp]
\centering
\caption{Receiver ablations on identical ScopeBind releases at frozen thresholds.}
\label{tab:ablation}
\SBMainTableFont

\begin{tabular*}{\linewidth}{@{\extracolsep{\fill}}l*{12}{c}@{}}
\toprule
\multirow[c]{2}{*}[-0.6ex]{Receiver} & \multicolumn{6}{c}{COCO ($n=160$)}
& \multicolumn{6}{c}{DiffusionDB ($n=160$)}\\
\cmidrule(lr){2-7}\cmidrule(lr){8-13}
& Strong$\uparrow$ & All$\uparrow$ & Crop$\uparrow$ & Copy FA$\downarrow$ & \SBHeaderStack{Unmarked\\FA$\downarrow$} & Wrong$\downarrow$ & Strong$\uparrow$ & All$\uparrow$ & Crop$\uparrow$ & Copy FA$\downarrow$ & \SBHeaderStack{Unmarked\\FA$\downarrow$} & Wrong$\downarrow$\\
\midrule
Original receiver & 119 & 1999 & 472 & 0 & 0 & 0 & 105 & 1969 & 464 & 0 & 0 & 0 \\
Joint verification & 136 & 2040 & 472 & 9 & 12 & 2 & 128 & 2017 & 465 & 7 & 0 & 0 \\
Joint + UID routing & 136 & 2028 & 472 & 0 & 12 & 2 & 127 & 2010 & 465 & 0 & 0 & 0 \\
Without gallery & 136 & 2028 & 472 & 0 & 0 & 0 & 127 & 2009 & 464 & 0 & 0 & 0 \\
One code proposal & 139 & 2033 & 472 & 0 & 0 & 0 & 128 & 2010 & 464 & 0 & 0 & 0 \\
SSCD--DINO agreement & 150 & 2045 & 472 & 0 & 0 & 0 & 139 & 2027 & 464 & 1 & 2 & 0 \\
\addlinespace[2pt]
\textbf{ScopeBind} & 153 & 2048 & 472 & 0 & 0 & 0 & 139 & 2027 & 464 & 1 & 2 & 0 \\
\bottomrule

\end{tabular*}
\par\smallskip
{\SBMainTableNote
Counts are reported per domain. Denominators are $160$ (Strong), $2{,}080$ (All), $480$ (Crop), $1{,}120$ (Copy FA), and $2{,}080$ (Unmarked FA and Wrong). The first four rows incrementally change verification. The following two rows replace only the single supplementary candidate.\par}
\end{table*}
}}
\par\smallskip
\SBPanel{\centering
\includegraphics[width=.80\linewidth]{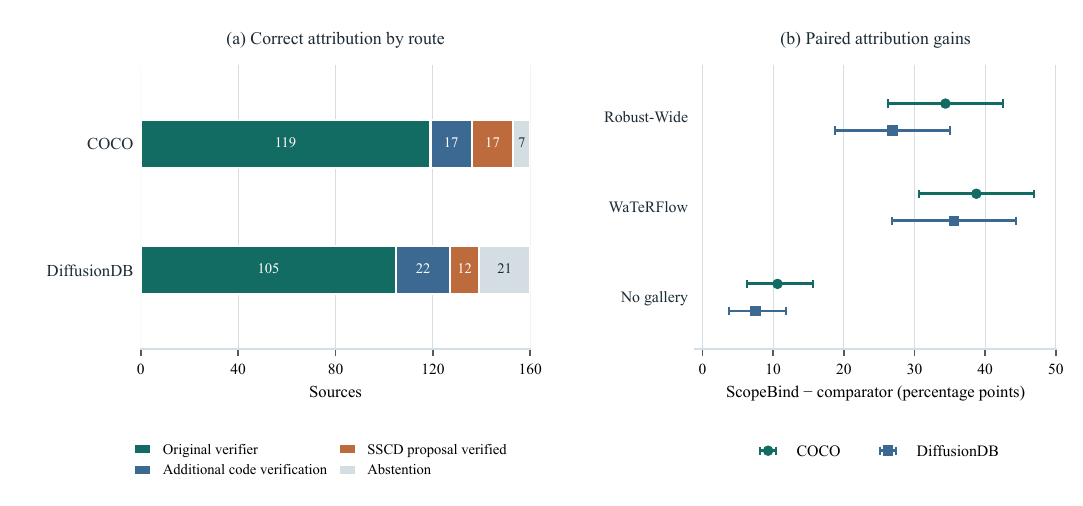}
\caption{Decision-route breakdowns and paired attribution gains at strength 0.5, counting the first route recovering each source alongside source-level bootstrap intervals.}\label{fig:stage_results}}
\par\smallskip
\SBTablePanel{\centering\caption{Controlled representation and candidate ablations on identical queries (counts per domain).}\label{tab:controlled}
\SBMainTableFont\begin{tabular*}{\linewidth}{@{\extracolsep{\fill}}l*{10}{r}@{}}
\toprule
\multirow[c]{2}{*}[-0.6ex]{Representation / control} & \multicolumn{5}{c}{COCO ($n=160$)} & \multicolumn{5}{c}{DiffusionDB ($n=160$)}\\
\cmidrule(lr){2-6}\cmidrule(lr){7-11}
& Strong$\uparrow$ & Crop$\uparrow$ & Copy FA$\downarrow$ & U FA$\downarrow$ & Wrong$\downarrow$ & Strong$\uparrow$ & Crop$\uparrow$ & Copy FA$\downarrow$ & U FA$\downarrow$ & Wrong$\downarrow$\\
\midrule
Endpoint distances (9D) & 121 & 472 & 3 & 35 & 0 & 105 & 464 & 5 & 51 & 0\\
Endpoint-derived geometry (7D) & 153 & 472 & 0 & 13 & 2 & 141 & 465 & 1 & 2 & 0\\
Projection only (4D) & 125 & 473 & 12 & 166 & 2 & 108 & 465 & 14 & 158 & 11\\
Remainder only (6D) & 119 & 472 & 2 & 23 & 0 & 114 & 464 & 7 & 50 & 0\\
Joint (10D) & 151 & 472 & 0 & 11 & 2 & 139 & 465 & 1 & 2 & 0\\
\textbf{Joint + content test ($g, h$)} & \textbf{153} & \textbf{472} & \textbf{0} & \textbf{0} & \textbf{0} & \textbf{139} & \textbf{464} & \textbf{1} & \textbf{2} & \textbf{0}\\
\midrule
Frozen ScopeBind (reallocated) & 153 & 472 & 0 & 0 & 0 & 139 & 464 & 1 & 2 & 0\\
Frozen, original candidate union & 154 & 472 & 0 & 0 & 0 & 139 & 464 & 1 & 2 & 0\\
\bottomrule
\end{tabular*}

\par\smallskip
{\SBMainTableNote Denominators per domain are 160 strong queries, 480 crops, 1,120 transfer controls, 2,080 unmarked controls (U FA), and 800 evaluated positives (clean, crop, and strong) for Wrong. The top six rows fit only the linear recovery head on identical 96-source fit and 32-source calibration splits using the same calibration rule; the bottom two retain the frozen head and thresholds, varying only candidate reallocation.\par}}
\par\smallskip

\ScopeColumns{To isolate the value of our orthogonal block-quartile representation beyond system-level routing, Table~\ref{tab:controlled} compares six recovery heads fitted and calibrated separately on identical registered inputs, frozen base decisions, and candidate lists: raw endpoint distances and norms ($\|e_s\|_2^2, \|e_s-d_s\|_2^2, \|d_s\|_2$ across $s\in\{1,4,8\}$, 9D), endpoint-derived geometry combining the four projection-related entries at $s\in\{1,4\}$ with three global orthogonal RMS terms $\log(1+\sqrt{\|v_s\|_2^2/m_s})$ across $s\in\{1,4,8\}$, where $m_s$ is the total number of latent entries across all channels and spatial positions at scale $s$ (7D), projection only (4D), block-remainder only (6D), the joint 10D representation, and the complete joint-plus-content rule. Raw endpoint distances recover only 121/160 COCO and 105/160 DiffusionDB strong queries while admitting 35 and 51 unmarked false acceptances. Projection alone admits 166 and 158 unmarked false acceptances (plus 12 and 14 copy false acceptances), while remainder alone stalls at 119 and 114 strong recoveries; both cues therefore play complementary roles under the evaluated protocol.}{Although algebraically recovering global orthogonal energy in the 7D control lifts strong recovery to 153 and 141, it still admits 13 unmarked false acceptances and two wrong UIDs on COCO. Pairing our 10D spatial-quartile representation with the explicit content test $h\ge\kappa$ eliminates all COCO false acceptances and wrong UIDs and achieves 153/160 strong recoveries with 0 copy false acceptances, 0 unmarked false acceptances, and 0 wrong UIDs. On DiffusionDB, it reproduces the aggregate performance of the frozen receiver (139/160 strong recoveries, 1 copy false acceptance, 2 unmarked false acceptances, and 0 wrong UIDs). Comparing the bottom two rows further clarifies the role of candidate reallocation. Without gallery fallback, reallocation raises code-mediated strong recovery from 121 to 136 (+15) on COCO and from 109 to 127 (+18) on DiffusionDB, substantially reducing dependence on visual retrieval. When the full gallery is active, both candidate sets reach virtually identical final accuracy (153 vs.\ 154 on COCO and 139 vs.\ 139 on DiffusionDB) with identical false-acceptance control.}

\SBNeedspace{50pt}
\subsection{Cross-Writer Generalization}\label{sec:writer_transfer}

\ScopeColumns{To test whether the orthogonal geometric verification core generalizes beyond the dual-carrier ScopeBind writer, we transfer the ten-feature construction to Robust-Wide, VINE, and the WaTeRFlow image codec. Each writer receives its own ternary linear softmax head fitted once on the same 96 sources and calibrated once on the same 32 separate sources as the original ScopeBind head. Each writer-specific head is fitted on nine positive and nine unmarked conditions per source alongside wrong-source and seven transfer controls, with no test source or cross-editor query entering fitting or calibration.}{Within each writer, all three receiver variants (regional K64, geometric-only, and geometric-plus-gallery) are evaluated on identical edited query images under a unified evaluation pipeline (consistent with Tables~\ref{tab:comparison}--\ref{tab:ablation} within a single query). The geometric variants replace regional acceptance with the two geometric criteria over each writer's original deduplicated top-64 candidates without inheriting unconditional K64 acceptances, while the full variant adds at most one eligible SSCD proposal after abstention across all 13 positive, 13 unmarked, and seven transfer conditions.}

\par\smallskip
\SBTablePanel{\centering\caption{Writer-specific registered verification on two held-out domains.}\label{tab:writer_generalization}
{\fontsize{7.4}{8.5}\selectfont\setlength{\tabcolsep}{3.2pt}\renewcommand{\arraystretch}{0.96}\begin{tabular*}{\linewidth}{@{\extracolsep{\fill}}l*{12}{c}@{}}
\toprule
\multirow[c]{2}{*}[-0.6ex]{Method} & \multicolumn{6}{c}{COCO ($n=160$)} & \multicolumn{6}{c}{DiffusionDB ($n=160$)}\\
\cmidrule(lr){2-7}\cmidrule(lr){8-13}
& Strong$\uparrow$ & All$\uparrow$ & Crop$\uparrow$ & Copy FA$\downarrow$ & U FA$\downarrow$ & \SBHeaderStack{Wrong\\UID$\downarrow$} & Strong$\uparrow$ & All$\uparrow$ & Crop$\uparrow$ & Copy FA$\downarrow$ & U FA$\downarrow$ & \SBHeaderStack{Wrong\\UID$\downarrow$}\\\midrule
Robust-Wide + K64 & 61.88 & \textbf{66.06} & \textbf{2.29} & \underline{1.61} & \underline{2.50} & \textbf{0.00} & \textbf{60.00} & \textbf{64.52} & \textbf{1.88} & \underline{0.63} & \underline{1.78} & \textbf{0.00}\\
Robust-Wide + Geometry & \underline{73.13} & 46.20 & \underline{0.00} & \textbf{0.00} & \textbf{0.00} & \textbf{0.00} & 49.38 & 42.74 & 0.00 & \textbf{0.00} & \textbf{0.05} & \textbf{0.00}\\
Robust-Wide + Geometry + SSCD & \textbf{86.88} & \underline{47.60} & \underline{0.00} & \textbf{0.00} & \textbf{0.00} & \textbf{0.00} & \underline{54.38} & \underline{43.70} & \underline{0.42} & \textbf{0.00} & \textbf{0.05} & \textbf{0.00}\\
\addlinespace[2pt]
VINE + K64 & \textbf{60.63} & \textbf{65.72} & \textbf{2.08} & \underline{1.25} & \underline{2.21} & \textbf{0.00} & \textbf{55.00} & \textbf{64.57} & \textbf{2.08} & 0.98 & 1.44 & \textbf{0.00}\\
VINE + Geometry & 27.50 & 29.04 & \underline{0.00} & \textbf{0.00} & \textbf{0.00} & \textbf{0.00} & 13.75 & 23.27 & 0.00 & \textbf{0.18} & \textbf{0.00} & \textbf{0.00}\\
VINE + Geometry + SSCD & \underline{30.63} & \underline{29.47} & \underline{0.00} & \textbf{0.00} & \textbf{0.00} & \textbf{0.00} & \underline{14.38} & \underline{23.75} & \underline{1.67} & \underline{0.45} & \underline{0.77} & \textbf{0.00}\\
\addlinespace[2pt]
WaTeRFlow + K64 & 56.88 & \textbf{79.57} & \textbf{31.88} & \underline{1.07} & 0.77 & \textbf{0.00} & \underline{51.25} & \textbf{77.31} & \textbf{29.79} & 1.16 & \underline{2.12} & \textbf{0.00}\\
WaTeRFlow + Geometry & \underline{73.13} & 47.21 & \underline{0.00} & \textbf{0.00} & \textbf{0.00} & \textbf{0.00} & 46.88 & 44.13 & 0.42 & \textbf{0.09} & \textbf{0.00} & \textbf{0.00}\\
WaTeRFlow + Geometry + SSCD & \textbf{82.50} & \underline{48.27} & \underline{0.00} & \textbf{0.00} & \underline{0.05} & \textbf{0.00} & \textbf{51.88} & \underline{44.71} & \underline{0.63} & \underline{0.18} & \textbf{0.00} & \textbf{0.00}\\
\addlinespace[2pt]
\bottomrule
\end{tabular*}
}
\par\smallskip
{\SBMainTableNote Entries are percentages; U FA denotes unmarked false acceptance. Bold and underlining identify the best and second-best distinct rates within each writer and domain.\par}}
\par\smallskip

\ScopeColumns{Relative to each writer's K64-adapted baseline (Table~\ref{tab:writer_generalization}), replacing regional acceptance with geometric verification and conditional retrieval improves strong-reconstruction attribution on COCO by +25.00 percentage points for Robust-Wide (to 86.88\%) and +25.63 points for WaTeRFlow (to 82.50\%), while holding false acceptances to at most 0.05\%. On DiffusionDB, WaTeRFlow improves slightly (+0.63 points to 51.88\%), whereas Robust-Wide decreases by 5.63 points (to 54.38\%) and VINE drops on both domains ($-30.00$ points on COCO and $-40.63$ points on DiffusionDB) under the conservative zero-calibration-FA threshold rule.}{Crop accuracy also drops when regional acceptance is replaced by global-only geometric scoring, which confirms that the dual-carrier local branch (Tables~\ref{tab:comparison} and~\ref{tab:nomination}) remains essential for cropped queries.}

\SBNeedspace{50pt}
\subsection{Manifold-Aligned Carrier Fusion}

\ScopeColumns{We next evaluate the finite angle-search writing rule in~\eqref{eq:aligned_write}--\eqref{eq:aligned_objective} against equal-weight fusion on the same 160-source domains. The equal-weight direction is included as a candidate for every image, and both writing rules are evaluated with the same fixed ScopeBind receiver, thresholds, and unmarked controls to ensure a fair comparison.}{}

\par\smallskip
\SBTablePanel{\centering\caption{Paired comparison of equal-weight and manifold-aligned writing with the frozen ScopeBind receiver.}\label{tab:aligned_fusion}
\SBMainTableFont\begin{tabular*}{\linewidth}{@{\extracolsep{\fill}}lcccccccc@{}}
\toprule
Method & Strong$\uparrow$ & All$\uparrow$ & Crop$\uparrow$ & Copy FA$\downarrow$ & U FA$\downarrow$ & Wrong UID$\downarrow$ & PSNR$\uparrow$ & SSIM$\uparrow$ \\
\midrule
\multicolumn{9}{c}{\textbf{COCO} ($n=160$)}\\\midrule
Equal-weight fusion & \underline{95.63} & \textbf{98.46} & \underline{98.33} & \textbf{0.00} & \textbf{0.00} & \textbf{0.00} & \textbf{35.06} & \textbf{0.9869} \\
Manifold-aligned fusion & \textbf{97.50} & \underline{98.41} & \textbf{99.58} & \textbf{0.00} & \textbf{0.00} & \textbf{0.00} & \underline{35.02} & \underline{0.9825} \\
\addlinespace[2pt]
\multicolumn{9}{c}{\textbf{DiffusionDB} ($n=160$)}\\\midrule
Equal-weight fusion & \underline{86.88} & \underline{97.40} & \underline{96.67} & \textbf{0.09} & \textbf{0.10} & \textbf{0.00} & \textbf{35.04} & \textbf{0.9883} \\
Manifold-aligned fusion & \textbf{89.38} & \textbf{97.55} & \textbf{98.13} & \textbf{0.09} & \underline{0.14} & \textbf{0.00} & \underline{35.03} & \underline{0.9853} \\
\addlinespace[2pt]
\bottomrule
\end{tabular*}

\par\smallskip
{\SBMainTableNote Attribution and FA entries (U FA: unmarked FA) are percentages; PSNR and SSIM are publication means, and every image meets the 35~dB floor. Equal-weight DiffusionDB All (97.40\%, 2,026/2,080) follows the unified evaluation pipeline of Section~\ref{sec:writer_transfer}. Bold and underlining rank the two rules within each domain.\par}}
\par\smallskip

\ScopeColumns{Manifold-aligned writing consistently improves strong-reconstruction and cropping attribution in both domains while maintaining strict false-acceptance control (Table~\ref{tab:aligned_fusion}). Specifically, on COCO it raises strong-reconstruction attribution from 153/160 (95.63\%) to 156/160 (97.50\%; paired gain +1.88 points, 95\% interval 0.00 to 4.38, gaining 3 sources with no regressions) and crop attribution from 472 to 478/480, while all-condition attribution shifts from 2,048 to 2,047/2,080 with zero transfer (0/1,120) and zero unmarked (0/2,080) false acceptances and a slight mean SSIM shift from 0.9869 to 0.9825. On DiffusionDB, strong attribution increases from 139/160 (86.88\%) to 143/160 (89.38\%; paired gain +2.50 points, 95\% interval 0.63 to 5.00, gaining 4 sources with no regressions), all-condition attribution rises from 2,026 to 2,029/2,080, and crops improve from 464 to 471/480; transfer false acceptances remain 1/1,120 while unmarked false acceptances move from 2 to 3/2,080 and mean SSIM shifts from 0.9883 to 0.9853.}{Decomposing the newly recovered strong queries clarifies where these gains arise: on COCO, 2 are accepted via code-candidate geometry and 1 via the supplementary gallery; on DiffusionDB, 1 is accepted via code-candidate geometry and 3 via the gallery. Comparing receiver configurations between equal-weight and manifold-aligned releases under the unified pipeline shows that Dual + K64 moves from 90 to 88 on COCO (versus 91 in Table~\ref{tab:comparison} under the primary evaluation) and from 85 to 83 on DiffusionDB, while gallery-free geometric recovery moves from 136 to 134 on COCO and from 127 to 124 on DiffusionDB; in contrast, conditional gallery retrieval increases from 17 to 22 strong recoveries on COCO and from 12 to 19 on DiffusionDB. Inspecting the writing objective components shows that the finite search increases the local-code margin on average while slightly lowering the projection and global-code terms in both domains, reflecting a clear trade-off between strong-reconstruction/crop recovery and publication fidelity under a fixed 35~dB PSNR floor (Fig.~\ref{fig:aligned_fidelity}).}

\par\smallskip\SBPanel{\centering\includegraphics[width=.82\linewidth]{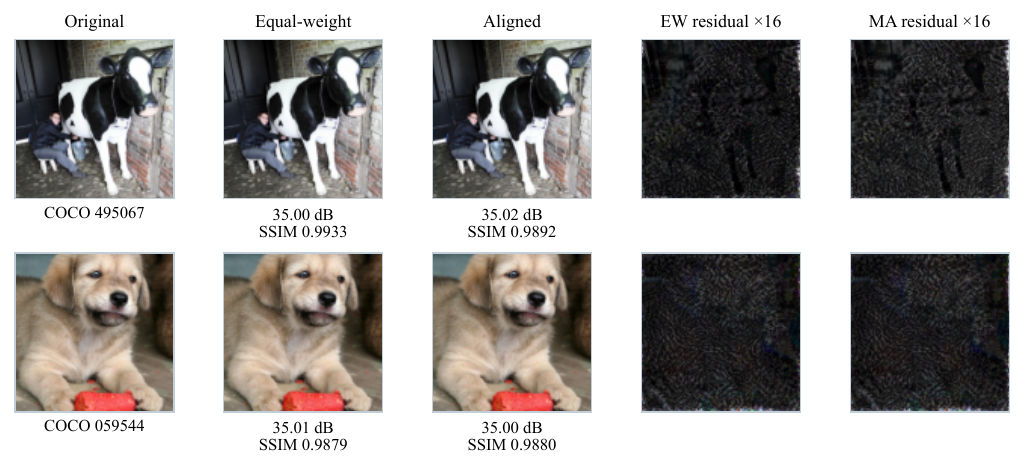}\caption{Original and watermarked releases for two representative COCO sources comparing equal-weight (EW) and manifold-aligned (MA) writing, with absolute RGB residuals amplified $16\times$ and measured PSNR/SSIM values.}\label{fig:aligned_fidelity}}
\par\smallskip

\SBNeedspace{50pt}
\subsection{Cross-Editor Generalization}

\ScopeColumns{To examine cross-editor behavior without refitting any parameter, we keep the SD2.1 verification VAE, both geometric thresholds, and all receiver heads frozen, and evaluate 160 COCO release queries and 160 unmarked queries under SDXL-base-1.0 (strengths 0.4 and 0.5, empty prompt, 50 steps, guidance 7.5) and InstructPix2Pix (fixed oil-painting instruction, 50 steps, text guidance 7.5, image guidance 1.5)~\cite{podell2023sdxl,brooks2023instruct}. All writers share source-specific seeds within each channel, and Dual and ScopeBind read identical release pixels. Although the editor VAEs share the AutoencoderKL architecture, they carry distinct trained weights (with distinct latent scaling factors) alongside different denoising backbones and schedulers.}{}

\par\smallskip
\SBTablePanel{\centering\caption{Cross-editor attribution with frozen receivers.}\label{tab:cross_editors}
\SBMainTableFont\begin{tabular*}{\linewidth}{@{\extracolsep{\fill}}l*{9}{c}@{}}
\toprule
\multirow[c]{2}{*}[-0.6ex]{Method} & \multicolumn{3}{c}{SDXL ($s=0.4$)} & \multicolumn{3}{c}{SDXL ($s=0.5$)} & \multicolumn{3}{c}{InstructPix2Pix (oil)}\\
\cmidrule(lr){2-4}\cmidrule(lr){5-7}\cmidrule(lr){8-10}
& Accuracy$\uparrow$ & U FA$\downarrow$ & Wrong UID$\downarrow$ & Accuracy$\uparrow$ & U FA$\downarrow$ & Wrong UID$\downarrow$ & Accuracy$\uparrow$ & U FA$\downarrow$ & Wrong UID$\downarrow$\\\midrule
Robust-Wide + K64 & 43.13 & 1.25 & \textbf{0} & 7.50 & \textbf{0.00} & \textbf{0} & 86.25 & 2.50 & \textbf{0}\\
Robust-Wide + Geometry + SSCD & 0.00 & \textbf{0.00} & \textbf{0} & 0.63 & \textbf{0.00} & \textbf{0} & 1.25 & \textbf{0.00} & \textbf{0}\\
VINE + K64 & 53.75 & \underline{0.63} & \textbf{0} & 8.13 & \underline{1.25} & \textbf{0} & 85.63 & 1.88 & \textbf{0}\\
VINE + Geometry + SSCD & 1.88 & \textbf{0.00} & \textbf{0} & 1.25 & \textbf{0.00} & \textbf{0} & 0.00 & \textbf{0.00} & \textbf{0}\\
Dual EW + K64 & 48.75 & 1.25 & \textbf{0} & 9.38 & 1.88 & \textbf{0} & 85.00 & 3.13 & \textbf{0}\\
ScopeBind (equal-weight) & \textbf{60.63} & \underline{0.63} & \textbf{0} & \textbf{18.13} & \textbf{0.00} & \textbf{0} & \textbf{89.38} & \underline{0.63} & \textbf{0}\\
Dual MA + K64 & 49.38 & 1.25 & \textbf{0} & 6.88 & 1.88 & \textbf{0} & 85.00 & 3.13 & \textbf{0}\\
ScopeBind (aligned) & \underline{56.25} & \underline{0.63} & \textbf{0} & \underline{14.38} & \textbf{0.00} & \textbf{0} & \underline{86.88} & \underline{0.63} & \textbf{0}\\
\bottomrule
\end{tabular*}

\par\smallskip
{\SBMainTableNote Accuracy and unmarked FA (U FA) are percentages; Wrong UID is a count among 160 positives; EW and MA denote equal-weight and manifold-aligned writing. Single-writer geometry variants use the frozen writer-specific heads from Section~\ref{sec:writer_transfer} on identical query pixels. Bold and underlining rank distinct scores within each editor.\par}}
\par\smallskip

\ScopeColumns{Across all three unseen editing channels, equal-weight ScopeBind achieves the highest accuracy among all compared methods (Table~\ref{tab:cross_editors}): 97/160 (60.63\%) on SDXL ($s=0.4$) versus 86/160 for VINE + K64, 29/160 (18.13\%) on SDXL ($s=0.5$) versus 15/160 for Dual EW + K64, and 143/160 (89.38\%) on InstructPix2Pix (oil painting) versus 138/160 for Robust-Wide + K64. Under these editors, it incurs at most one false acceptance among 160 unmarked queries (0/160 on SDXL $s=0.5$) and zero wrong UIDs. Manifold-aligned writing trails equal-weight writing across all three unseen channels (90/160, 23/160, and 139/160, respectively); this indicates that optimizing against the SD2.1 reconstruction proxy favors the matched PNDM channel rather than guaranteeing editor-independent gains. Likewise, the frozen single-writer geometric heads exhibit very low acceptance under these unseen editors due to the calibration sensitivity of writer-specific heads under editing-distribution shifts (Fig.~\ref{fig:cross_editor_examples}).}{}

\par\smallskip\SBPanel{\centering\includegraphics[width=.79\linewidth]{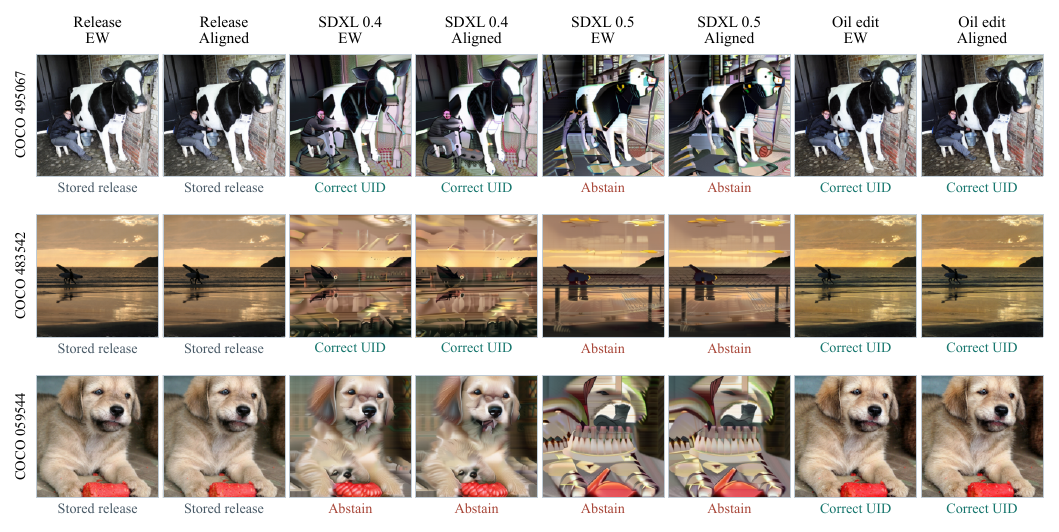}\caption{Three representative natural-image examples under the cross-editor protocol. EW denotes equal-weight writing; each label indicates the receiver decision on the edited query.}\label{fig:cross_editor_examples}}
\par\smallskip

\SBNeedspace{50pt}
\subsection{Gallery Scaling and Computational Cost}

\ScopeColumns{We evaluate retrieval scaling across nested original-only galleries containing the 160 issued sources augmented with unissued COCO distractors to reach total gallery sizes $N\in\{256, 1{,}000, 5{,}000, 10{,}000\}$. Only the SSCD search pool expands; issued records, blind candidate lists, UID occupancy gating, and both geometric thresholds remain fixed. Because the added distractors are unissued originals, this stress test isolates retrieval recall stability as gallery size grows rather than modeling 10,000 concurrently issued UIDs.}{As the gallery scales 39-fold from 256 to 10,000 originals, aligned strong-query SSCD top-1 retrieval remains 160/160 on COCO and shifts only from 160/160 to 159/160 on DiffusionDB (Fig.~\ref{fig:gallery_scaling}). Final strong attribution remains unchanged at 156/160 on COCO (0/1,120 transfer FA, 0/2,080 unmarked FA) and 143/160 on DiffusionDB (1/1,120 transfer FA, 3/2,080 unmarked FA).}

\par\smallskip\SBPanel{\centering\includegraphics[width=.56\linewidth]{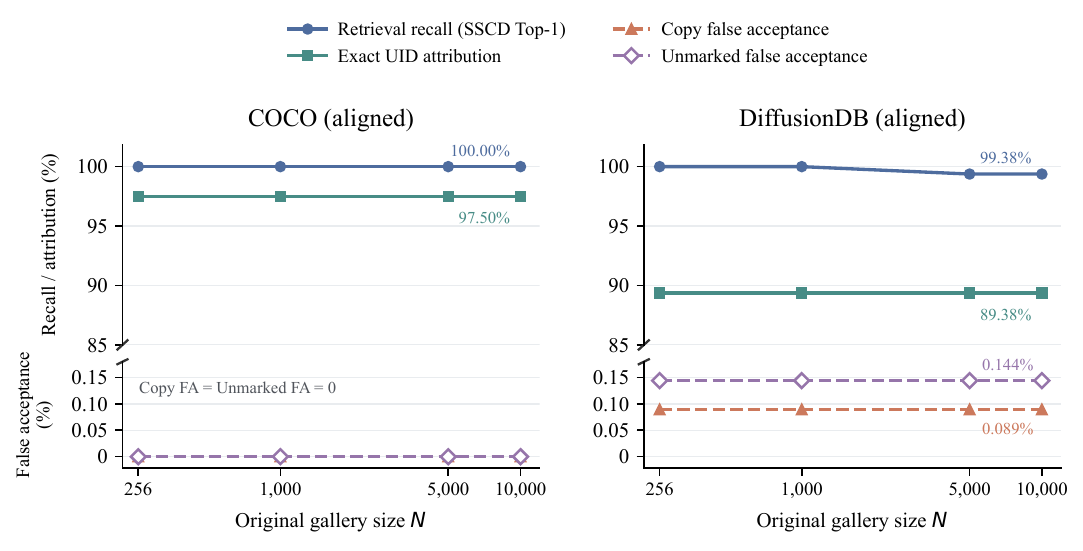}\caption{Nested original-only gallery scaling at a fixed receiver operating point. Upper panels show strong-query SSCD top-1 retrieval recall and exact UID attribution; lower panels show false acceptance across negative controls.}\label{fig:gallery_scaling}}
\par\smallskip

\begin{multicols*}{2}
\SBTablePanel{\centering\caption{Inference latency on RTX 4090 and per-record registration storage.}\label{tab:efficiency}
{\fontsize{7.3}{8.4}\selectfont\setlength{\tabcolsep}{3pt}\renewcommand{\arraystretch}{0.95}\renewcommand{\arraystretch}{1.02}
\begin{tabular*}{\linewidth}{@{\extracolsep{\fill}}lrrr@{}}
\toprule
\multicolumn{4}{l}{\textbf{(a) Inference Latency (RTX 4090)}} \\
\midrule
Operation & Med.\ (ms) & P95 (ms) & $n$ \\
\midrule
Equal-weight writing & 332.2 & 431.5 & 36 \\
Manifold-aligned writing & 5464.3 & 5799.9 & 36 \\
Global \& local (19 views) & 333.0 & 339.9 & 36 \\
Query VAE encoding & 22.0 & 22.8 & 36 \\
Geometry (per candidate) & 1.33 & 1.41 & 36 \\
SSCD feature extraction & 5.47 & 5.74 & 24 \\
FAISS IP (10k, CPU) & 0.70 & 0.75 & 1000 \\
Complete receiver, base acc. & 1238.0 & 1490.7 & 6 \\
Complete receiver, code rec. & 1161.8 & 1472.4 & 6 \\
Complete receiver, gallery rec. & 995.4 & 1030.8 & 6 \\
Complete receiver, abstention & 1131.8 & 1497.6 & 6 \\
\midrule
\multicolumn{4}{l}{\textbf{(b) Per-Record Registration Storage (KiB)}} \\
\midrule
\multicolumn{3}{l}{Paired VAE means (FP16)} & 64.00 \\
\multicolumn{3}{l}{SSCD descriptor (FP32)} & 2.00 \\
\multicolumn{3}{l}{\textbf{Incremental cache subtotal}} & \textbf{66.00} \\
\multicolumn{3}{l}{Regional tokens} & 192.00 \\
\multicolumn{3}{l}{Regional grayscale reference} & 49.00 \\
\multicolumn{3}{l}{Original \& release images (median)} & 898.90 \\
\bottomrule
\end{tabular*}
}
\par\smallskip
{\SBMainTableNote Shared neural-model weights are amortized separately. Complete receiver rows report end-to-end per-query latencies across all four decision routes; FAISS IP excludes SSCD extraction (5.47~ms).\par}}
\par\smallskip

Exact inner-product retrieval matches FAISS IndexFlatIP~\cite{douze2024faiss}. Table~\ref{tab:efficiency} reports component and end-to-end inference latencies on a single RTX~4090 GPU. Equal-weight writing takes 332.2~ms (median), and the 18-candidate manifold-aligned search takes 5.46~s. On the verification side, 19-view global and local readout requires 333.0~ms, query VAE encoding takes 22.0~ms, each geometric candidate check takes 1.33~ms, query SSCD feature extraction takes 5.47~ms, and exact 10,000-record CPU FAISS search takes 0.70~ms. Complete end-to-end receiver latency remains around 1.0--1.2~s across all four decision paths (1,238.0~ms base acceptance, 1,161.8~ms code recovery, 995.4~ms gallery recovery, and 1,131.8~ms abstention). Per issued record, the paired FP16 VAE means (64~KiB) and FP32 SSCD descriptor (2~KiB) add 66~KiB of incremental geometric-and-search cache, alongside the regional tokens (192~KiB), grayscale reference (49~KiB), and original and release images (898.9~KiB median); original-only distractors store only their compact descriptors and identifiers.
\par\smallskip

\section{Discussion}
The registered pair is most valuable precisely when visual similarity and watermark decoding disagree: an archived original can reintroduce the right source into contention, yet a strong visual match to the unmarked original is equally compatible with an unauthorized source-only query. A concise information-theoretic bound makes this trade-off explicit. For a fixed UID and editing channel, let $P_u$ and $Q_u$ denote the query distributions induced by the issued release and its unmarked original, respectively. For any acceptance event $A_u$ with the registry fixed,
\begin{equation}
P_u(A_u)-Q_u(A_u)\leq \mathrm{TV}(P_u,Q_u),
\end{equation}
so any channel that drives the two distributions together fundamentally limits the separation between true acceptance and same-source false acceptance. ScopeBind's orthogonal geometric test exploits whatever release displacement survives editing, while our writer-specific, manifold-aligned, cross-editor, and large-gallery experiments map out where that surviving separation remains reliable.

Several practical boundaries remain open for future study. First, the finite manifold-aligned search optimizes a deterministic SD2.1 reconstruction proxy and improves matched-channel robustness, whereas equal-weight fusion transfers better across unseen editors such as SDXL and InstructPix2Pix. Second, our large-gallery stress tests expand unissued original-only distractors up to 10,000 records while holding the 160 issued releases fixed; scaling to tens of thousands of concurrently occupied UIDs or disambiguating multiple separately watermarked releases of the same original raises distinct collision and gating challenges. Finally, our negative controls cover a fixed set of nonadaptive transplantation attacks and a single instruction-editing prompt; evaluating registered-release verification against adaptive adversaries targeting the full verifier remains an important open problem.

\section{Conclusion}
ScopeBind separates visual similarity to an archived original from verifiable lineage to a registered watermarked release. By deriving a release-specific issuance direction from each registered original--release pair and decomposing query displacements into parallel issuance-alignment and orthogonal content-consistency components, a single fitted verification head evaluates bounded global and local watermark hypotheses alongside one conditional gallery proposal. Controlled representation ablations show that combining issuance alignment with explicit orthogonal content testing eliminates the false-acceptance vulnerabilities of raw endpoint distances and single-component features, while the constrained manifold-aligned writing rule strengthens matched-channel robustness without retraining the underlying codecs. Evaluations across natural and generated image domains, same-pixel controls, writer-specific verification transfer, unseen generative editors (SDXL and InstructPix2Pix), and galleries scaled to 10,000 originals establish both the attribution gains and the operating boundaries of this approach. Extending registered-release verification to adaptive transplantation adversaries, multiple watermarked releases per original, and large-scale concurrently occupied UID spaces remains an important direction for future work.

\par

\end{multicols*}
\end{document}